\documentclass[11pt]{article}
\usepackage{eamt26}
\usepackage{times}
\usepackage{url}
\usepackage{latexsym}
\usepackage[small,bf]{caption} 
\usepackage[hidelinks]{hyperref}
\makeatletter
\AtBeginDocument{%
  \def\@lbibitem[#1]#2{\item[]\if@filesw
      {\begingroup
        \let\protect\noexpand
        \immediate\write\@auxout{\string\bibcite{#2}{#1}}%
      \endgroup}\fi
      \Hy@raisedlink{\hyper@anchorstart{cite.#2\@extra@b@citeb}\hyper@anchorend}%
      \ignorespaces}%
}
\makeatother
\title{AI-assisted cultural heritage dissemination: Comparing NMT and glossary-augmented LLM translation in rock art documents}

\author{Vicent Briva-Iglesias\\
  SALIS, CTTS, ADAPT Centre\\
  Dublin City University\\
  {\tt vicent.brivaiglesias@dcu.ie}
  \And
  Maria Ferre-Fernández\\
  Universidad de Almería\\
  {\tt mff181@ual.es}
}
\date{}

\begin{document}
\maketitle
\begin{abstract}
  Cultural heritage institutions increasingly disseminate research and interpretive materials globally, but multilingual dissemination is constrained by limited budgets and staffing. In terminology-dense domains such as rock art, translation quality depends on accurate, consistent specialised terms, and small lexical errors can mislead non-specialists and reduce reuse. We compare three English MT setups for a Spanish academic rock art text, focusing on simple, operationally feasible interventions rather than complex model-side modifications: (1) DeepL as a strong NMT baseline, (2) Gemini-Simple (LLM with a basic prompt), and (3) Gemini-RAG (the same LLM with glossary-augmented prompting via term-pair retrieval). Using PEARMUT, we conduct a human evaluation via (i) multi-way Direct Assessment (0--100) and (ii) targeted terminology auditing with a restricted MQM taxonomy. Gemini-RAG yields the highest exact-match terminology accuracy (81.4\%), versus Gemini-Simple (69.1\%) and DeepL (64.4\%), while preserving overall quality (mean DA 85.3 Gemini-RAG vs. 85.2 Gemini-Simple), outperforming DeepL (80.3). These results show that glossary-augmented prompting is a low-overhead way to improve terminology control in cultural-heritage translation if institutions maintain minimal terminology resources and lightweight evaluation procedures.
\end{abstract}

\section{Introduction}

Digital infrastructures have expanded the reach of cultural heritage scholarship and interpretation, but multilingual access remains uneven. Large platforms (e.g., Europeana and related initiatives) have explicitly explored machine translation (MT) as a route to scaling multilingual access to heritage metadata and content, reflecting institutional pressures to broaden accessibility without proportionate growth in translation budgets \cite{kaldeliEuropeanaTranslateProviding2022}. Cultural heritage also appears in global sustainability agendas, including SDG Target 11.4 (“protect and safeguard the world’s cultural and natural heritage”), which reinforces the societal value of dissemination and accessibility \cite{SDGIndicator1141,pettiCulturalHeritageSustainable2020}.

In this context, rock art dissemination is a particularly demanding challenge for MT. Rock art documentation and interpretation rely on specialised vocabulary \cite{domingoLatestDevelopmentsRock2013,valdez-tullettDigitalRockArt2023}. Within rock art dissemination, translation errors may constitute a critical problem: a mistranslated motif label or an inconsistent rendering of a chronocultural category can distort interpretation and reduce trust, especially when translations are reused for education, outreach, or indexing.

The recent rise of large language models (LLMs) has shifted the MT landscape \cite{brownLanguageModelsAre2020}. LLMs often produce fluent translations under simple instructions \cite{gaoHowDesignTranslation2023,jiaoChatGPTGoodTranslator2023}, but professional acceptance frequently depends on control of these technologies - especially terminology fidelity and consistency in specialised domains (see, for example, Briva-Iglesias et al. \shortcite{briva-iglesiasLargeLanguageModels2024} in legal translation). Terminology control has a long history in MT research (e.g., lexically constrained decoding for NMT) \cite{haslerNeuralMachineTranslation2018,postFastLexicallyConstrained2018}. 

For LLMs, “control” often takes the form of prompting strategies or augmentation with external resources such as dictionaries or glossaries \cite{kimEfficientTerminologyIntegration2024}. A practical and increasingly common approach is retrieval-augmented generation (RAG), in which relevant context is retrieved and injected into the prompt to steer output \cite{lewisRetrievalAugmentedGenerationKnowledgeIntensive2020}. This paper has two overarching research questions relevant to AI-assisted heritage dissemination workflows:

\begin{itemize}
  \item RQ1. How do an LLM baseline and a glossary-augmented LLM compare to a strong commercial NMT baseline for overall translation quality of a terminology-dense rock art text?
  \item RQ2. Does lightweight glossary augmentation measurably improve terminology accuracy (exact-match to preferred English forms) and reduce terminology error types (wrong/missing/inconsistent), as judged by professional annotators?
\end{itemize} 
To answer these questions, we conduct a small-scale human evaluation using PEARMUT~\cite{zouharPearmutHumanEvaluation2026}, combining multi-way direct ment (DA)-style quality ratings with targeted terminology evaluation under a restricted MQM taxonomy. We then interpret results through the lens of deployability: what minimal resources are sufficient to produce meaningful gains in terminology control for cultural heritage dissemination via AI-powered language technologies.

\section{Background and related work}
Specialised translation requires both the transfer of vocabulary from one language to another and the mediation of domain-specific knowledge through linguistically and conceptually appropriate forms. Across terminology studies and translation studies, terminology is commonly treated as the organising principle of specialised discourse and, by extension, a central component of specialised translation \cite{cabreicastellviTerminologyCommunication1999,cabreicastellviTraductorTerminologiaNecesidad2000,martinezTerminologicalCompetenceTranslation2009,scarpaIntroducingSpecialisedTranslation2020}. Terms are embedded in conceptual systems rather than functioning as isolated lexical units, which means that translation problems in specialised domains often arise from mismatches between knowledge structures, disciplinary conventions, and preferred usage rather than from language alone \cite{cabreicastellviTraductorTerminologiaNecesidad2000,faberbenitezTerminologySpecializedLanguage2012,maksymenkoFeaturesTranslatingScientific2023}.

These concerns become especially acute in cultural heritage. Like highly standardised scientific and technical domains, cultural heritage communication often combines specialist description, interpretation, institutional mediation, and public-facing dissemination. However, translation in heritage contexts is also shaped by practical constraints such as cost, time, and spatial limitations, especially where multilingual provision must fit fixed label formats or platform-specific requirements \cite{ghaziTranslationPracticesMuseums2022,liaoTranslatingMultimodalTexts2018}. At the same time, heritage institutions increasingly need to disseminate content across languages at scale. Europeana Translate is a clear example of this tendency, having explored MT as a way of increasing multilingual access to cultural heritage resources \cite{kaldeliEuropeanaTranslateProviding2022}. 

Other terminological initiatives in cultural heritage protection and documentation, including resources associated with FISH and Getty, as well as broader work on AI and cultural heritage protection, show that structured terminologies are already recognised as essential infrastructures for description, documentation, and access \cite{colaceNewAIChallenges2025,forumoninformationstandardsinheritageTerminologies2024,forumoninformationstandardsinheritageFISHTerminologies2026,gettyresearchinstituteCulturalObjectsName2017,gettyresearchinstituteArtArchitectureThesaurus2021}. However, these resources are often fragmented, unevenly multilingual, or not easily operationalised within translation workflows. As a result, institutions frequently rely on “good-enough” and risk-managed multilingual dissemination strategies rather than fully standardised end-to-end solutions \cite{kaldeliEuropeanaTranslateProviding2022}.

Rock art provides a particularly revealing test case within this broader heritage landscape. Rock art scholarship depends on descriptive terminology for motifs, techniques, surfaces, and recording practices, but also on interpretive categories and chronocultural labels that may be historically layered, theoretically contested, and shaped by local research traditions \cite{whitley-item-4001,mazelRockArtDating2007}. This makes terminology in rock art unusually sensitive for translation. Recording and analysis in the field increasingly rely on digital methods and enhancement tools, and digital archaeology has further expanded the visibility and reuse of rock art documentation in research and dissemination contexts \cite{domingoLatestDevelopmentsRock2013,valdez-tullettDigitalRockArt2023}. In such settings, terminology errors are not trivial: they can misrepresent archaeological content, confuse non-specialist readers, and weaken indexing and retrieval across repositories and heritage platforms \cite{mason2006}. The problem is compounded by the fact that rock art terminology is not fully stable even within the field itself. Chippindale~\shortcite{chippindaleWhatAreRight2001} highlighted the lack of standardised terminology and even questioned the use of the term art for certain markings, while Mazel, Nash, and Waddington~\shortcite{mazelRockArtDating2007} similarly point to the absence of international consensus on key lexical units. This instability makes terminological support resources especially valuable. Existing glossaries and reference resources illustrate the field’s ongoing effort to consolidate and clarify terminology for both professional and broader audiences (Bednarik, 2003, 2010, 2026; Bradshaw Foundation, n.d.; Research Laboratories of Archaeology, n.d.; Sabo and Sabo, 2006; Scottish Rock Art Project, 2021).

In this context, evaluating MT for rock art dissemination requires more than a general assessment of fluency. It requires explicit attention to terminology control. In NMT, terminology constraints have been studied extensively, particularly through lexically constrained decoding, which shows that enforcing user-specified terms is possible but not trivial \cite{haslerNeuralMachineTranslation2018,postFastLexicallyConstrained2018}. In LLM-based translation, control is more often implemented through prompting and augmentation with external lexical resources than through decoding-level constraints \cite{gaoHowDesignTranslation2023}. Recent work suggests that dictionary- and glossary-based augmentation can improve the translation of rare or specialised items by injecting structured lexical guidance into the prompt \cite{kimEfficientTerminologyIntegration2024}. Retrieval-augmented generation (RAG) provides a broader framework for this type of intervention, allowing relevant external information to be retrieved dynamically and supplied at generation time without retraining the model \cite{lewisRetrievalAugmentedGenerationKnowledgeIntensive2020}. For terminology-sensitive heritage workflows, this is particularly attractive because it offers a lightweight and operationally feasible way to increase lexical control.

The question then becomes how such gains should be evaluated, provided that evaluation of translation quality is a complex issue \cite{rossiHowChooseSuitable2022}, and in such specialised domains the evaluation only becomes more complicated. Human evaluation remains the most informative approach for MT quality assessment \cite{laubliSetRecommendationsAssessing2020}, even if practical constraints often encourage over-reliance on automatic metrics \cite{hanMachineTranslationEvaluation2018}. Direct Assessment (DA) has been widely used to capture overall translation quality through continuous human judgments \cite{grahamAccurateEvaluationSegmentlevel2015}, while MQM offers a structured framework for diagnosing specific error types, including terminology-related problems \cite{kocmiFindingsWmt25General2025}. More recently, PEARMUT has been proposed as a lightweight platform for implementing DA-, error span annotation, and MQM-style evaluation protocols with lower setup overhead (Zouhar and Kocmi, 2026). For the present study, this combination is especially relevant. In a domain such as rock art, a translation can be globally fluent and adequate while still failing to follow preferred terminology. A methodology that combines holistic assessment (DA-based evaluation) with targeted terminology auditing (MQM-based evaluation) is therefore better suited to the actual problem than either approach alone. This is precisely the gap addressed in the present paper: not whether AI systems can produce fluent English translations of heritage texts, but whether lightweight glossary augmentation can improve terminology control in a domain where lexical precision, consistency, and interpretive trust are crucial.

\section{Materials}
\subsection{Source text and glossary}
The source material is a Spanish academic rock art text divided into 91 segments, totalling 1,743 Spanish words. This text is a fragment from a published paper in Rock Art Research \cite{delaralopezChronoculturalProposalAtlanterra2025}. The text is a complex document that contains specialised rock art terminology. The text was segmented using sentence-level segmentation aligned with punctuation in the original publication: segments are short enough for reliable comparative judgement while preserving local discourse coherence.

We also use a bilingual glossary of 200 Spanish-English preferred term pairs as the terminology resource for glossary-augmented prompting and for terminology evaluation. For the targeted terminology evaluation, we restrict analysis to glossary terms that actually appear in the source text: 44 distinct expected English terms, with 194 total term occurrences across the 91 segments. We also add non-relevant terms to the glossary to add noise to the retrieval of the content and assess how the MT systems perform. This glossary was created by one of the authors following the recommendations of good practices in the revised material.

\subsection{Systems compared}
We compare three English MT configurations for the same Spanish segments:
First, we use DeepL as the NMT baseline. It is a commercial NMT system accessed via API at the time of translation, on March 2026, selecting the “Classic Language model” (NMT) as opposed to the “Next-gen language model” (LLM) \cite{DeepLTranslatorWorlds}.

Second, we use gemini-3.1-pro-preview in a configuration that we call “Gemini-Simple”. This is our LLM baseline and was accessed via API and a temperature of 1. While lower temperatures (e.g., 0 or 0.2) are traditionally favoured to maximize determinism, we intentionally retained the default temperature of 1. This decision serves as a robust stress test for the RAG intervention: if we give the model its full generative variance, we evaluate whether lightweight prompt augmentation is strong enough to override the LLM’s inherent lexical fluidity. Gemini-Simple uses a commercial LLM system with a minimal prompt (“Translate the following text from Spanish to English”), without explicit terminology guidance, and using the recommendation by Jiao et al. \shortcite{jiaoChatGPTGoodTranslator2023}, ranking no. 3 in LMArena at the time of writing, March 2026 \cite{chiangChatbotArenaOpen2024}. This indicates that it is a frontier model.

Third, we use gemini-3.1-pro-preview in a configuration that we call “Gemini-RAG”. This is our glossary-augmented LLM and was accessed via API and a temperature of 1. This system uses the same LLM as “Gemini-Simple", but we add prompt augmentation via lightweight retrieval of relevant glossary entries. For each segment, we retrieve glossary entries whose Spanish term appears in the segment (case-insensitive exact string match, allowing simple punctuation boundaries). Retrieved entries are formatted as explicit constraints (“Use the preferred English term exactly as written; keep consistent across the text”). This is a deliberately lightweight “RAG-inspired” operationalisation: retrieval is deterministic and transparent, and augmentation is achieved by injecting the term list into the prompt rather than modifying decoding or retraining.
This strategy is motivated by two strands of literature: (i) RAG as a general mechanism for injecting external knowledge into generation \cite{lewisRetrievalAugmentedGenerationKnowledgeIntensive2020}, and (ii) evidence that dictionary/glossary augmentation can improve translation performance on difficult lexical items \cite{kimEfficientTerminologyIntegration2024}.

\section{Method}
\subsection{Human evaluation design in PEARMUT}
Human evaluation was conducted in PEARMUT using two complementary tasks. The first task targets overall translation quality through DA-style scoring. The second targets terminology compliance through a restricted MQM-style audit. The rationale for combining these two tasks is straightforward. If the study relied only on overall quality, it might miss terminology failures that do not strongly affect surface readability. If it relied only on terminology evaluation, it would say little about whether the translations remain globally acceptable as English outputs in a specialised translation domain. The combination of both tasks therefore reflects the dual nature of the research problem and allows us to respond to the overarching RQs.

\paragraph{Task 1: Direct Assessment:}
For the first task, two annotators were shown the Spanish source segment together with the three candidate English MT proposals side by side (see Figure 1). System identities were anonymised and output order was randomised by segment. Annotators assigned a score from 0 to 100 to each candidate for overall translation quality. In this study, overall quality was understood holistically, combining meaning preservation with readability and appropriateness for academic dissemination.

\begin{figure*}[t]
  \centering
  \setlength{\fboxsep}{0pt}
  \fbox{\includegraphics[width=0.92\textwidth]{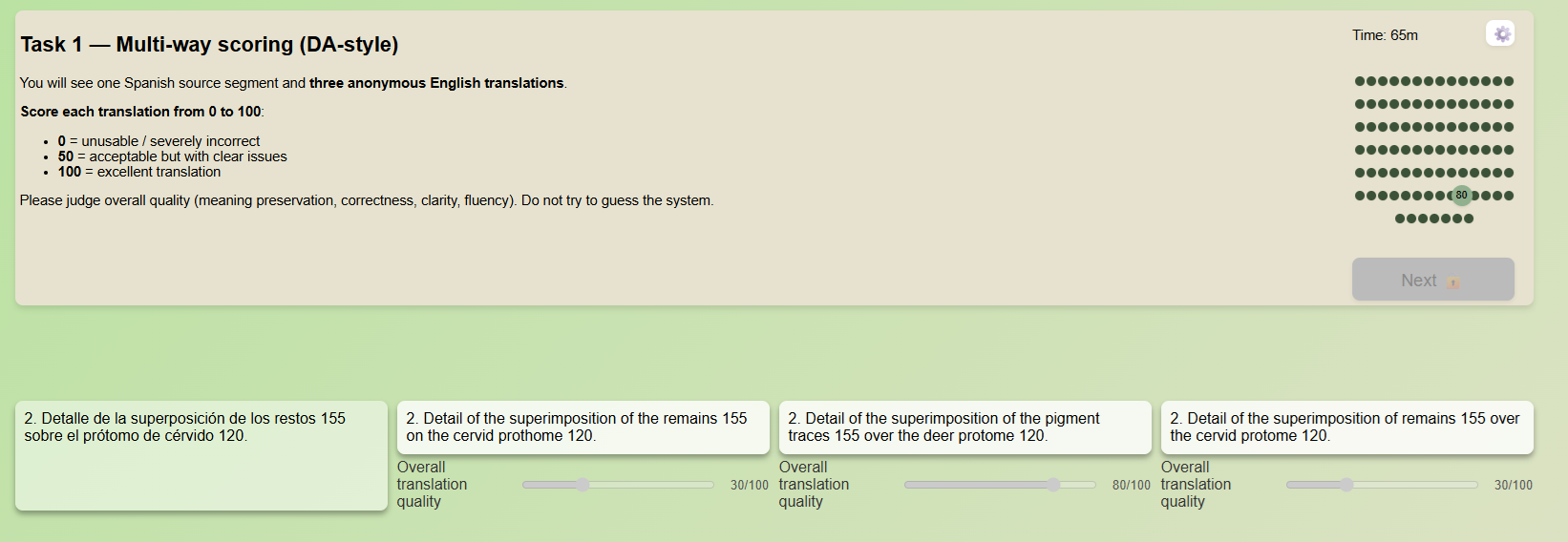}}
  \caption{PEARMUT interface for Task~1 (multi-way DA-style quality rating). For each Spanish source segment, three anonymised system outputs are shown side by side and scored on a 0--100 scale.}
  \label{fig:task1}
\end{figure*}

\begin{figure*}[t]
  \centering
  \setlength{\fboxsep}{0pt}
  \fbox{\includegraphics[width=0.92\textwidth]{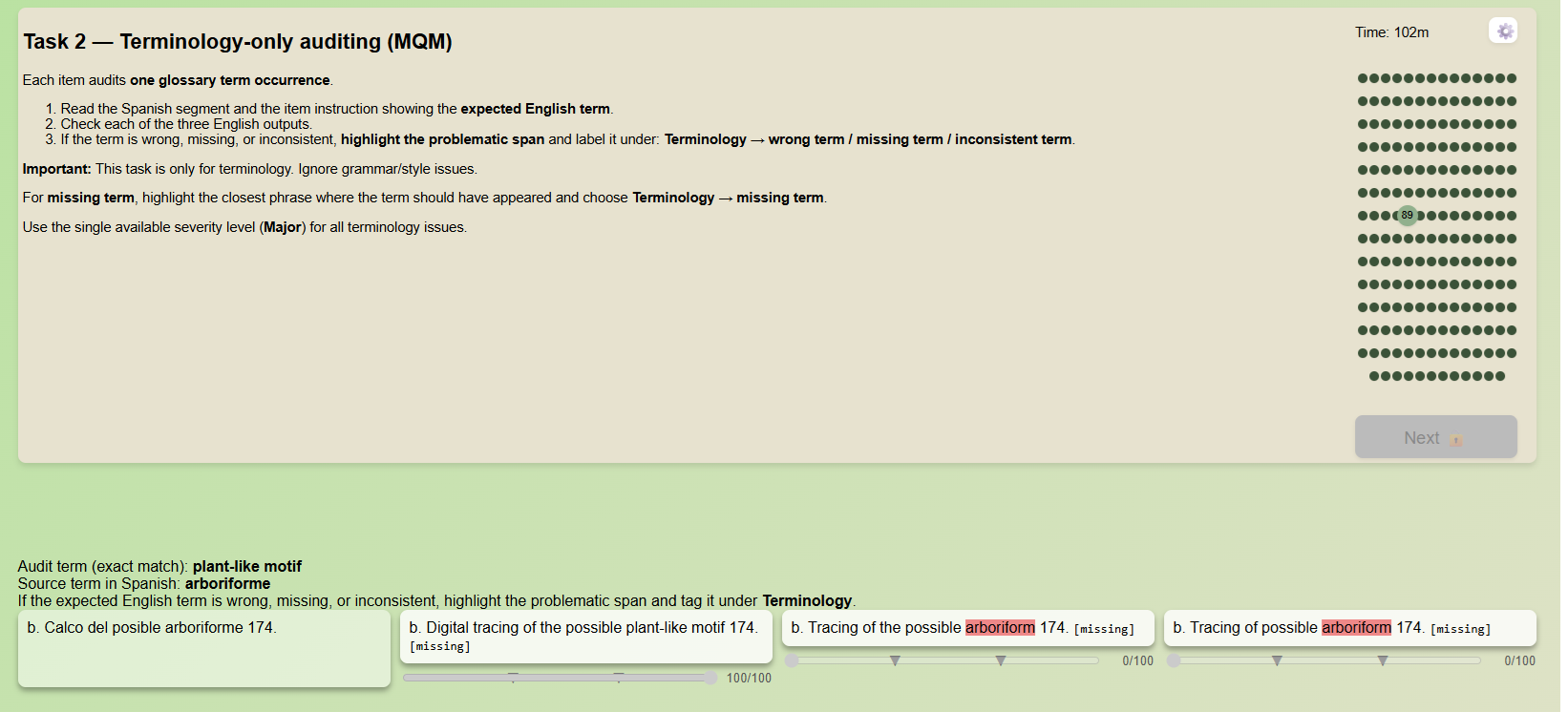}}
  \caption{PEARMUT interface for Task~2 (targeted terminology audit). Each item shows the Spanish source segment, the expected glossary term, and three anonymised system outputs for labeling terminology errors as wrong, missing, or inconsistent.}
  \label{fig:task2}
\end{figure*}

Strictly speaking, this is DA-style rather than classical DA in its original standalone form, because the outputs are seen in comparison rather than in isolation. The contrastive interface was used intentionally. When systems are relatively close, side-by-side presentation can improve sensitivity to nuanced differences while retaining the advantages of continuous scoring. Across 91 segments and three systems, this task produced 273 segment-system ratings. 

For analysis, the paper reports mean DA scores by system, paired per-segment score differences, 95\% bootstrap confidence intervals based on 5,000 resamples over segments, and Wilcoxon signed-rank tests as exploratory inferential support. These statistics are not presented as proof of universal system superiority, but as structured ways of describing the comparative behaviour of the systems on this specific dataset.

\paragraph{Task 2: Terminology Evaluation:}
The second task implemented a terminology-only MQM-style protocol. Instead of applying a full MQM taxonomy across all dimensions of quality \cite{mqmcoreMQMMultidimensionalQuality2025}, the study restricts the analysis to “No error" or three terminology-related labels: wrong term, missing term, and inconsistent term. Each audit item displays the Spanish source segment, the expected English term from the glossary, and the outputs from the three systems. Annotators mark a system output when the preferred target form is not respected and assign the relevant terminology error label, if applicable (see Figure 2).

\begin{table*}[t]
  \centering
  \small
  \caption{DA-style overall quality summary over 91 evaluated segments. Top performing configuration in bold.}
  \label{tab:da-summary}
  \begin{tabular}{lccc}
    \hline
    System & Mean DA score & Standard deviation & Segments ($n$) \\
    \hline
    DeepL & 80.27 & 19.34 & 91 \\
    Gemini-Simple & 85.24 & 16.00 & 91 \\
    Gemini-RAG & \textbf{85.27} & 19.05 & 91 \\
    \hline
  \end{tabular}
\end{table*}

\begin{table*}[t]
  \centering
  \small
  \caption{Paired segment-level DA comparisons (91 paired segments; bootstrap CIs over segments). Asterisk for statistical significance.}
  \label{tab:da-paired}
  \begin{tabular}{lccc}
    \hline
    Pairwise contrast & Mean DA difference & 95\% bootstrap CI & Wilcoxon $p$ \\
    \hline
    Gemini-RAG $-$ DeepL & +5.00 & [0.27, 9.40] & .0078* \\
    Gemini-Simple $-$ DeepL & +4.97 & [1.65, 8.64] & .0020* \\
    Gemini-RAG $-$ Gemini-Simple & +0.03 & [$-$4.74, 4.38] & .324 \\
    \hline
  \end{tabular}
\end{table*}

This design significantly reduces annotation burden while preserving diagnostic relevance for the paper’s main RQs. Full-scale error annotation is often unrealistic in small studies involving specialist translators. By contrast, term-level auditing can be focused, efficient, and closely aligned with domain-specific quality needs. The fact that the system also guides the annotator and indicates what term to expect helps in the costly human evaluation. The evaluation covers 194 term occurrences across the three systems, producing 582 segment-system term checks. Exact-match terminology accuracy is operationalised in the following way: an item counts as correct only when the expected preferred English form appears exactly as specified. 

Because the same term occurrence is evaluated across all three MT systems, system comparisons can be made within item. The paper therefore uses exact McNemar tests on correct versus incorrect outcomes as an exploratory way of comparing terminology adherence across systems.

\subsection{Annotators and adjudication}
One professional annotator with +5 years of professional experience in specialised translation completed the full evaluation across both tasks. A second professional annotator with +10 years of professional experience in language and translation technologies then reviewed the completed annotations, after which both annotators discussed questionable cases and aligned final decisions through adjudication. This process was intended to improve consistency and reduce annotation drift, especially for terminology items where the boundary between an exact preferred form and a plausible but non-preferred alternative can be important.

The design is not equivalent to a full independent double annotation procedure. The study does not report inter-annotator agreement because it does not include two separate primary annotation layers. The final judgments are adjudicated expert decisions, not consensus statistics derived from parallel annotation. The paper therefore treats the annotation design as a strength in terms of careful review, but also as a limitation in terms of formal evaluator robustness \cite{artsteinInterannotatorAgreement2017}.

\section{Results}
\subsection{Overall translation quality}
Across the 91 evaluated segments, the two Gemini conditions received almost identical mean DA-style scores, and both outperformed the NMT baseline on this dataset. As shown in Table 1, mean DA scores were 85.27 for Gemini-RAG, 85.24 for Gemini-Simple, and 80.27 for DeepL. The corresponding standard deviations were 19.05, 16.00, and 19.34, respectively, indicating a degree of segment-level variability across all systems that is typical of small, domain-specific evaluation sets.

Paired per-segment comparisons, summarised in Table 2 point in the same direction. Gemini-RAG exceeded DeepL by an average of 5.00 DA points (95\% bootstrap CI [0.27, 9.40]; Wilcoxon p=.0078), and Gemini-Simple exceeded DeepL by 4.97 points (95\% CI [1.65, 8.64]; Wilcoxon p=.0020), with statistically significant differences. By contrast, the difference between the two Gemini conditions in DA scores was minimal and not significant (+0.03 points; 95\% CI [$-$4.74, 4.38]; Wilcoxon p=.324). These results suggest that, for this terminology-dense rock art text, both LLM configurations were judged more favourably than the NMT baseline in terms of overall perceived quality, while glossary augmentation did not materially alter the holistic acceptability of the LLM output in DA scores.

This pattern is consistent with broader observations in the recent MT and LLM literature: current LLM-based translation outputs are often perceived as highly fluent and readable under relatively simple prompting conditions, even when the main practical challenges lie elsewhere, such as in lexical control, consistency, or domain alignment \cite{kimEfficientTerminologyIntegration2024,brownLanguageModelsAre2020,briva-iglesiasLargeLanguageModels2024}. The present results therefore suggest that the principal value of glossary augmentation in this study does not lie in improving already strong surface-level quality, but in strengthening terminology governance without degrading overall translation quality, as described below.

\begin{table*}[t]
  \centering
  \small
  \caption{Exact-match terminology accuracy over 194 audited term occurrences per system. Top performing configuration in bold.}
  \label{tab:term-accuracy}
  \begin{tabular}{lccc}
    \hline
    System & Correct terms & Accuracy (\%) & Audited terms ($n$) \\
    \hline
    DeepL & 125/194 & 64.43 & 194 \\
    Gemini-Simple & 134/194 & 69.07 & 194 \\
    Gemini-RAG & \textbf{158/194} & \textbf{81.44} & 194 \\
    \hline
  \end{tabular}
\end{table*}

\begin{table*}[t]
  \centering
  \small
  \caption{Pairwise exact McNemar tests for terminology correctness. Asterisk for statistical significance.}
  \label{tab:term-mcnemar}
  \begin{tabular}{lc}
    \hline
    Pairwise contrast & Exact McNemar $p$ \\
    \hline
    Gemini-RAG vs. DeepL & $<$.00001* \\
    Gemini-RAG vs. Gemini-Simple & $<$.001* \\
    Gemini-Simple vs. DeepL & .064 \\
    \hline
  \end{tabular}
\end{table*}

\subsection{Exact-match terminology accuracy}
Terminology results showed a clearer separation between systems than the overall quality scores. Over the 194 audited term occurrences, exact-match terminology accuracy was highest for Gemini-RAG, which achieved 81.44\% correctness (158/194), followed by Gemini-Simple with 69.07\% (134/194) and DeepL with 64.43\% (125/194), as presented in Table 3. These findings indicate that the most pronounced empirical advantage in the study lies in terminology adherence rather than in general fluency or adequacy.

Pairwise within-item comparisons reinforce this interpretation. As shown in Table 4, Gemini-RAG significantly outperformed both baselines under exact McNemar testing: Gemini-RAG vs. DeepL: p = .00001; Gemini-RAG vs. Gemini-Simple: p = .001. By contrast, the difference between Gemini-Simple and DeepL did not reach statistically significance (p = .064). In other words, the non-augmented LLM baseline was not clearly superior to the NMT baseline in strict terminology compliance, whereas the glossary-augmented LLM was.

This is a key result for the paper’s overall argument. Prior work on terminology integration in both NMT and LLM-based translation has shown that lexical control remains a persistent difficulty and often requires explicit intervention, whether through constrained decoding, dictionary augmentation, or prompt-based lexical steering \cite{haslerNeuralMachineTranslation2018,postFastLexicallyConstrained2018,kimEfficientTerminologyIntegration2024}. The present case study supports that literature from a specialised translation perspective in general, and a cultural heritage perspective in particular: a lightweight glossary-augmentation strategy was sufficient to produce a substantial gain in exact preferred-term adherence without requiring retraining or more complex decoding methods.

\subsection{Terminology error profile}
A more fine-grained view of terminology behaviour emerges from the error-type distribution. Counting annotated terminology spans under the restricted MQM taxonomy yielded interesting results worth discussing (Table 5).

\begin{table*}[t]
  \centering
  \small
  \caption{Terminology error profile under the restricted MQM taxonomy. Top performing configuration in bold. Lower is better}
  \label{tab:term-errors}
  \begin{tabular}{lcccc}
    \hline
    System & Wrong term & Inconsistent term & Missing term & Total error spans \\
    \hline
    DeepL & 40 & 37 & \textbf{0} & 77 \\
    Gemini-Simple & 37 & 31 & \textbf{0} & 68 \\
    Gemini-RAG & \textbf{20} & \textbf{14} & 6 & \textbf{40} \\
    \hline
  \end{tabular}
\end{table*}
Two patterns are especially noteworthy. First, “wrong term" and “inconsistent term" dominate the error profile of the baseline systems. This suggests that, when no explicit terminology guidance is supplied, the systems frequently rely on uncontrolled lexical choice or oscillate across competing English renderings for conceptually related items. This is supported by recent research on contextual issues of MT \cite{castilhoSurveyContextNeural2025}. Second, Gemini-RAG substantially reduces both wrong-term and inconsistency errors, but introduces a small number of “missing term" cases. This indicates that the glossary-augmented system is more disciplined overall, yet still occasionally avoids the preferred form through paraphrase or omission, which is penalised under the strict exact-match definition used in this study.

This shift in the error profile is significant. In practical post-editing terms, research supports that repeatedly normalising inconsistent or systematically non-preferred lexical choices across a document set is often more burdensome than dealing with a smaller number of isolated omissions \cite{briva-iglesiasFosteringHumancenteredAugmented2024}. From that perspective, the contribution of glossary augmentation is not only higher accuracy, but a potentially more manageable error landscape for human reviewers. This is in line with broader translator-centred arguments for terminology-aware AI workflows, where the goal is not merely better output, but output that is easier to verify, standardise, and maintain across institutional materials \cite{scarpaIntroducingSpecialisedTranslation2020,forisRoleDocumentationDocument2021}.

\subsection{Illustrative qualitative examples} 
The terminology evaluation surfaced several recurring cases in which Gemini-RAG aligned output more closely with preferred domain terminology than Gemini-Simple and DeepL. One representative example concerns the Spanish term “pinturas rupestres", for which the preferred English form in the glossary was “rock paintings". In the baseline conditions, the systems frequently produced “cave paintings", which is plausible in general English but not equivalent in all archaeological or heritage contexts and may imply a narrower spatial setting. Gemini-RAG was much more likely to follow the preferred form, thereby aligning more closely with the project’s terminological policy.

At the same time, the qualitative evaluation also revealed the limits of the exact-match evaluation. In a smaller number of instances, Gemini-RAG generated a fluent paraphrase or reformulation that avoided the exact preferred target form. Under the study’s evaluation protocol, such cases were counted as incorrect, even when the broader meaning remained acceptable. This illustrates both the strength and the strictness of the terminology-oriented metric: it is well suited to assessing compliance with a defined term list, but it is narrower than a general measure of semantic adequacy. 

These examples are important because they ground the quantitative findings in domain-relevant translation behaviour. They show why terminology-sensitive evaluation cannot be reduced to a generic fluency-adequacy judgment, especially in cultural heritage and archaeology, where lexical choices often carry conceptual and interpretive weight \cite{chippindaleWhatAreRight2001,mazelRockArtDating2007,bednarikRockArtGlossary2010,bednarikIFRAOGlossary}.

\section{Discussion}
\subsection{Main findings}
This study provides evidence that lightweight glossary augmentation and a simple and easily deployable approach can yield substantial gains in terminology adherence without sacrificing perceived overall translation quality in a terminology-dense rock art dissemination task. The present results are noteworthy in showing that even lightweight interventions can produce substantial improvements in terminology control. As shown in Tables 1–5, the two Gemini conditions were almost indistinguishable in overall quality, yet clearly separated in exact-match terminology performance, with Gemini-RAG outperforming both Gemini-Simple and DeepL. 

The central implication is that glossary augmentation contributes primarily by improving terminological control, not by dramatically changing surface-level fluency. This distinction matters. Current LLMs already produce highly readable output under relatively simple prompts \cite{gaoHowDesignTranslation2023,jiaoChatGPTGoodTranslator2023,hendyHowGoodAre2023}, but specialised translation requires more than readability. Specialised translation requires alignment with preferred domain terminology and institutional practice \cite{briva-iglesiasAreAIAgents2025}. Prior work on terminology-constrained NMT and glossary-augmented LLM translation has repeatedly argued that lexical control is one of the key barriers to robust deployment in specialised domains \cite{haslerNeuralMachineTranslation2018,postFastLexicallyConstrained2018,kimEfficientTerminologyIntegration2024}. The present results support that argument in a cultural heritage setting, but opens the discussion of the applicability of RAG-augmented LLM translation in other specialised domains. 

Just as importantly, these gains were achieved through a low-overhead, transparent intervention: retrieve relevant glossary entries and inject them into the prompt. No fine-tuning, document-level retrieval pipeline, or specialised constrained-decoding infrastructure was required. In practical terms, this suggests that even modest terminology resources may already provide meaningful leverage when paired with LLM-based translation.

\subsection{Why DA alone is not enough for terminology-sensitive domains}
A key methodological takeaway is that overall quality evaluation and terminology compliance do not fully coincide. As the contrast between Tables 1 and 3 makes clear, Gemini-Simple and Gemini-RAG appear nearly identical if one looks only at mean DA scores, yet they differ substantially in terminology adherence. Segment-level inspection also showed cases where translations received high DA scores while still missing preferred terms. This divergence is not accidental, and it follows from the nature of the evaluation constructs themselves. DA is designed to capture perceived overall quality, whereas terminology auditing tests whether a translation conforms to an explicit lexical policy. 

For specialised translation in general, and professional heritage dissemination in particular, this distinction is crucial. A non-preferred term may still be fluent, plausible, and broadly adequate, which means it may escape strong penalty in an overall scoring task. Yet from an institutional perspective, that same term may still be undesirable because it undermines consistency, affects indexing, or departs from accepted disciplinary usage. MQM-style approaches are helpful precisely because they allow these dimensions to be separated analytically rather than collapsed into a single global score \cite{lommelMultidimensionalQualityMetrics2014}. 

The present study therefore supports a two-layer evaluation design for terminology-sensitive domains: one layer for broad overall quality (i.e. DA) and another for terminology-specific compliance. In small-scale studies and institutional pilot projects, this combination offers considerably more diagnostic value than either measure alone while remaining feasible within the lighter-weight evaluation environment enabled by PEARMUT~\cite{zouharPearmutHumanEvaluation2026}.

\subsection{Implications for cultural heritage institutions and translators}
From an operational perspective, the results support a pragmatic strategy for institutions seeking to scale multilingual dissemination under limited resources. First, institutions can benefit from creating or consolidating even a minimal glossary, prioritising high-impact lexical items such as motif labels, technique names, chronocultural categories, and conservation vocabulary. Second, glossary entries can be injected selectively through deterministic RAG rather than passed wholesale to the model, preserving simplicity and transparency. Third, quality evaluation can focus on terminology-sensitive points rather than requiring full-scale error annotation for every output. This workflow logic is consistent with the broader heritage context, where multilingual dissemination often has to balance accessibility, budget constraints, and institutional trust \cite{kaldeliEuropeanaTranslateProviding2022,ghaziTranslationPracticesMuseums2022,liaoTranslatingMultimodalTexts2018}.

For translators and domain experts, the proposed workflow is not a replacement for professional judgment. It could be understood as a form of human-centred AI language technology augmentation \cite{briva-iglesiasHumanCenteredAILanguage2026}. Glossary augmentation can reduce repeated term-hunting, promote lexical consistency, and give reviewers a more explicit basis for quality control, reducing repetitive corrections, which research shows increase cognitive load \cite{laubliTranslationTechnologyResearch}. This aligns with long-standing arguments in terminology and specialised translation research that reliable translation depends not only on linguistic competence, but also on documentation practices and explicit terminological management \cite{cabreicastellviTerminologiaRepresentacionComunicacion2010,scarpaIntroducingSpecialisedTranslation2020,forisRoleDocumentationDocument2021}.

More broadly, the study points toward a realistic model of AI-assisted heritage translation: not frictionless automation, but a risk-managed workflow in which lexical control is strengthened through compact resources and focused human review. In this respect, the findings are encouraging because they suggest that institutions do not need exhaustive multilingual termbanks before they can begin to benefit from terminology-aware AI-assisted dissemination. A small, ad-hoc 200-word glossary substantially improved terminological control in MT, even when using the same baseline LLM (Gemini-Simple vs Gemini-RAG).

\subsection{Limitations}
Several limitations frame the interpretation of the study. First, the dataset is small and narrow. It consists of a single Spanish rock art text segmented into 91 units. The findings therefore support a careful case-study claim, not a generalisable ranking across all heritage texts, all domains, or all language pairs. The field would benefit from replication across additional rock art materials, other archaeological subdomains, and different institutional text types. For the ease of replicability, the dataset and human evaluation scores are available in the following repository: \url{https://zenodo.org/records/20178898}.

Second, the annotation design is limited in scope. Although the study includes review and adjudication by a second professional annotator, it does not include a full independent parallel annotation layer due to budget constraints. This means that the final decisions are carefully reviewed expert judgments, but evaluator variability cannot be quantified through standard agreement measures, such as inter-annotator agreement. Future work should address this by including multiple independent annotators, if resources allow.

Third, the exact-match terminology metric is intentionally strict. That strictness is appropriate for testing adherence to a preferred term list, but it does not capture the full space of acceptable semantic alternatives. Some outputs counted as incorrect may still be acceptable under a looser policy. This is not a weakness of the paper so much as a reminder that evaluation metrics are always tied to operational goals. Here, the goal is compliance with preferred forms, not broad semantic permissibility.

Fourth, the systems studied are commercial services observed at a specific moment in time (dates of access and specific settings and models are provided in the paper). Model behaviour, interfaces, and hidden defaults may evolve. The results therefore describe comparative behaviour under the conditions of this study rather than stable properties of the systems across time.

\subsection{Future work}
There are several productive directions for extending this line of research. One is scale. Future studies should test the same design on larger and more varied datasets, including different rock art traditions, institutional genres, and language pairs. Another is annotation robustness. Independent multi-annotator evaluation would make it possible to estimate agreement and better understand how sensitive the findings are to evaluator variation \cite{artsteinInterannotatorAgreement2017}. An additional important direction is to test transfer to other specialised, terminology-heavy domains, like legal translation and healthcare.

Researching this in more naturalistic tasks would also be of relevance \cite{mellingerChapter2Designing2025}. It would be useful to move beyond output quality alone and examine post-editing effort directly. If glossary-augmented MT outputs reduce revision time or decrease the number of terminology corrections required, that would strengthen the practical case for deployment. Future work could also explore more contextual evaluation that distinguishes between exact preferred-form matches, acceptable variants, and conceptually correct paraphrases. This would be especially valuable in domains where institutional terminology policies are flexible rather than strict.

Finally, there is a governance dimension. Cultural heritage institutions adopting AI-assisted translation need not only technical tools, but also procedures for glossary maintenance, version control, policy documentation, and reviewer oversight. As AI becomes more embedded in heritage dissemination, these organisational questions will matter as much as model performance \cite{briva-iglesiasHumanCenteredAILanguage2026}. 

\section{Conclusion}
This paper examined terminology-sensitive AI-assisted translation for cultural heritage dissemination through a focused Spanish-English rock art case study. Comparing a commercial NMT baseline, a minimally prompted LLM baseline, and a glossary-augmented LLM condition, we found glossary augmentation produced the clearest benefit in exact-match terminology adherence while leaving perceived overall translation quality essentially unchanged relative to the non-augmented LLM. In other words, augmentation's main value in this study was not to make translations sound better in a broad sense, but to make them more lexically controllable in a domain where preferred terminology matters.

The study also makes a broader methodological point. In specialised dissemination contexts, overall quality scores do not necessarily imply terminological suitability. A translation may be fluent, adequate, and readable while still failing to follow the lexical policy required by a domain or institution. Therefore, evaluation designs that combine overall quality with targeted terminology auditing are especially valuable in specialised workflows.

The paper's broader contribution is therefore modest but practical. It shows that a small, explicit terminology resource can materially improve lexical control when paired with simple retrieval and prompt augmentation. For resource-constrained cultural heritage organisations, this offers a realistic way to strengthen multilingual dissemination without assuming that generic AI output alone is sufficient. Glossary-augmented prompting is not a universal solution, but it is a feasible, low-overhead control mechanism for terminology-sensitive cultural heritage dissemination.

\bibliography{eamt26}
\bibliographystyle{eamt26}
\end{document}